\documentclass[runningheads]{llncs}

\usepackage{eccv}

\usepackage{eccvabbrv}

\usepackage{graphicx}
\usepackage{booktabs}
\usepackage{placeins}
\usepackage{float}

\usepackage[accsupp]{axessibility}  

\usepackage{hyperref}

\usepackage{orcidlink}
\usepackage{amssymb}
\usepackage{amsmath}
\usepackage{amsfonts}

\usepackage{xcolor}
\usepackage{color}
\begin{document}

\title{Genetic Information Analysis of Age-Related Macular Degeneration Fellow Eye Using Multi-Modal Selective ViT} 

\titlerunning{Multi-Modal Selective ViT for Genetic Information Analysis}

\author{Yoichi Furukawa\inst{1}\orcidlink{0009-0004-2739-9901} \and
Satoshi Kamiya\inst{2}\orcidlink{0000-0002-7057-3280} \and
Yoichi Sakurada\inst{3}\orcidlink{0000-0002-2894-0454} \and
Kenji Kashiwagi\inst{3}\orcidlink{0000-0001-8506-8503} \and
Kazuhiro Hotta\inst{1}\orcidlink{0000-0002-5675-8713}}

\authorrunning{Y. Furukawa et al.}

\institute{Meijo University \and
Mitsubishi Electric Advanced Technology R\&D Center \and
Yamanashi University 
\email{243427034@ccmailg.meijo-u.ac.jp, 180442042@ccalumni.meijo-u.ac.jp, sakurada@yamanashi.ac.jp, kenjik@yamanashi.ac.jp, kazuhotta@meijo-u.ac.jp}}

\maketitle

\begin{abstract}
 In recent years, there has been significant development in the analysis of medical data using machine learning. It is believed that the onset of Age-related Macular Degeneration (AMD) is associated with genetic polymorphisms. However, genetic analysis is costly, and artificial intelligence may offer assistance. This paper presents a method that predict the presence of multiple susceptibility genes for AMD using fundus and Optical Coherence Tomography (OCT) images, as well as medical records. Experimental results demonstrate that integrating information from multiple modalities can effectively predict the presence of susceptibility genes with over 80$\%$ accuracy.

\end{abstract}

\section{Introduction}
\label{sec:intro}

In recent years, the usage of machine learning for data analysis in the medical field has advanced significantly\cite{litjens2017survey}. The onset of Age-related Macular Degeneration (AMD) is strongly associated with genetic polymorphisms, but genetic analysis is expensive and raises ethical concerns. Typically, specialists diagnose AMD using fundus images, OCT images, and other medical records. However, conventional models \cite{huang2020fusion} have not fully accounted for the relationships between these different types of images and tabular data.

This paper aims to effectively handle multiple medical images and non-image data such as medical records to predict the number of risk alleles in disease-related genes. Specifically, to predict samples with a risk allele number of 2 in the ARMS2 and CFH genes, which are susceptibility genes related to the eye disease AMD (age-related macular degeneration), it comprehensively utilizes fundus images, OCT images, and medical record information

Unlike the conventional ViT \cite{dosovitskiy2020image}, which inputs only a single image, our proposed method embeds fundus and OCT images using different patch embeddings and medical records using a Multilayer Perceptron (MLP) into the same dimensional space. We then introduce selective attention, and more selective and detailed information using a CNN.
Additionally, to reduce the asymmetry in information content between image and non-image data, we concurrently train to reconstruct the medical records and apply TSIA to resolve the mismatch in the number of fundus and OCT images, thus aiming to improve accuracy.

In experiments, we use 1,192 sets of fundus images, corresponding OCT images, zero-value pseudo images, and medical records (age, gender, smoking history).
We demonstrated that it is possible to predict the presence of risk alleles number 2 in the ARMS2 and CFH genes with accuracies of over 80\%.

The structure of this paper is as follows.
Section 2 discusses related researches.
Section 3 describes the details of the proposed method.
Section 4 explains the dataset and experimental results.
Section 5 presents the results of an ablation study.
Section 6 concludes with discussions on future challenges.

\section{Related Works}

In recent years, the Vision Transformer (ViT) \cite{dosovitskiy2020image} has achieved success  in various tasks such as classification, segmentation, and object detection   \cite{parmar2018image,bello2019attention,beal2020toward}. using the Transformer \cite{vaswani2017attention} architecture. ViT segments images into patches, converts them into vectors, adds a class token for classification, and incorporates trainable positional embeddings to consider the context of positions. This is particularly important in medical imaging, where distant objects often influence each other. Models like TransUnet \cite{chen2021transunet} and SwinUnet \cite{cao2022swin} utilize these relationships for segmentation. Additionally, traditional CNNs\cite{he2016deep,simonyan2014very} struggle to easily handle relationships between multiple modalities, such as images and text. 
In this paper, we apply ViT to treat images and non-images
as tokens and construct a model that considers their relationships through the attention mechanism.
Moreover, when we apply deep learning with multiple modalities to the medical field, it often happens that the number of images does not match across modalities. This discrepancy can introduce bias during training. Therefore, we propose a Table-based Similar Image Augmentation(TSIA),
where during training, the number of images per modality for each patient is supplemented with images from another patient whose medical record information has the highest cosine similarity to the non-image data of the target patient.

\section{Proposed method}
\subsection{Multi-modal Selective ViT (MSViT)}

\begin{figure}[ht]
        \centering
        \includegraphics[width=0.7\linewidth]{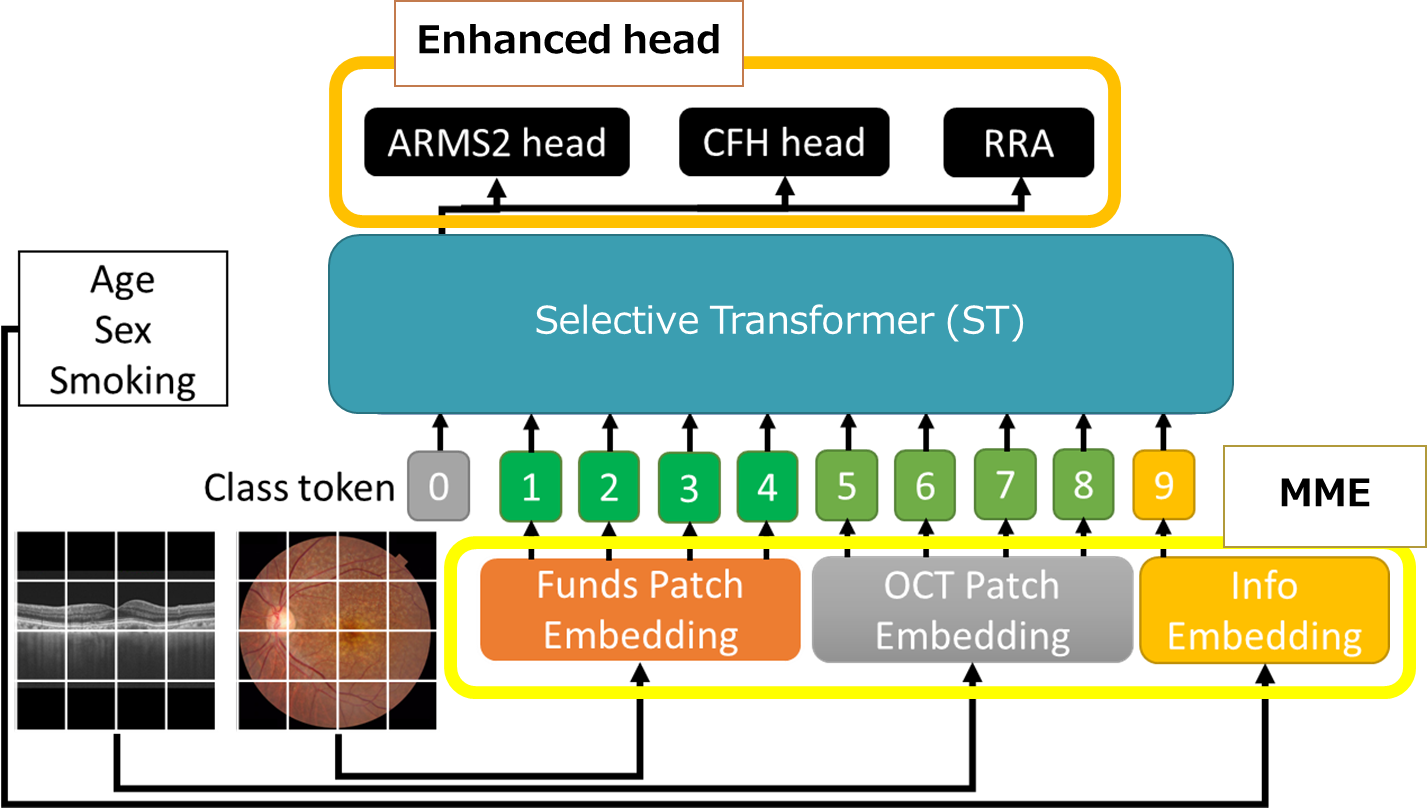}
        \caption{To process images and text simultaneously, MSViT includes a Multi-Modal Embedding (MME) for embedding information into tokens, a Selective Transformer (ST) with selective attention to tokens based on learned probabilities and dense feature extraction using a CNN, and an Enhanced head for classification.}
        \label{fig:MSViT}
    \end{figure}

To simultaneously process images and non-images, we propose the MSViT as shown in Figure \ref{fig:MSViT}. MSViT consists of a Multi-Modal Embedding (MME) that embeds information not only from different images\cite{xu2021deepchange} but also from table data into tokens,a Selective Transformer (ST) selective attention to tokens based on learned probabilities and dense feature extraction using a CNN. It further enhances information extraction by using a CNN on selected image patches, integrating detailed local features with global token information. This ensures the model focuses on the most informative regions, boosting overall effectiveness. An Enhanced head is head for classification, strengthened by learning to reconstruct table information. The modules are described in detail in the following sections.

\subsubsection{Multi-Modal Embedding (MME)}

Fundus images and OCT images are denoted as $I_{\text{Fundus}} \in \mathbb{R}^{H \times W \times 3}$ and $I_{\text{OCT}} \in \mathbb{R}^{H \times W \times 1}$ respectively.
We use separate patch embedding processes for two modalities. 
For fundus images,
\begin{equation}
    I_{\text{Fundus}} = \{I_{\text{Fundus}}^i \in \mathbb{R}^{P \times P \times 3} \mid i = 1, 2, \ldots, \frac{HW}{P^2} \}
\end{equation}
where $I_{\text{Fundus}}^i$ represents the $i$-th patch in the fundus image, and $P \times P$ is the size of each patch.
The patches are then embedded into tokens as
\begin{equation}
    z_{\text{Fundus}}^i = f(I_{\text{Fundus}}^i) \in \mathbb{R}^D \quad \forall i
\end{equation}
where $f(\cdot)$ is a linear projection function, and $z_{\text{Fundus}}^i$ is the embedded token for the $i$-th patch.
Similarly, for OCT images,
\begin{equation}
    I_{\text{OCT}} = \{I_{\text{OCT}}^j \in \mathbb{R}^{P \times P \times 1} \mid j = 1, 2, \ldots, \frac{HW}{P^2} \}
\end{equation}
where $I_{\text{OCT}}^j$ represents the $j$-th patch in the OCT image.
The patches are then embedded into tokens as
\begin{equation}
    z_{\text{OCT}}^j = g(I_{\text{OCT}}^j) \in \mathbb{R}^D \quad \forall j
\end{equation}
where $g(\cdot)$ is a linear projection function, and $z_{\text{OCT}}^j$ is the embedded token for the $j$-th patch.
The resulting patch tokens for fundus and OCT images are
\begin{equation}
    z_{\text{Fundus}} \in \mathbb{R}^{(HW/P^2) \times D}, \quad z_{\text{OCT}} \in \mathbb{R}^{(HW/P^2) \times D}.
\end{equation}

Next, the embeddings of the two different modality images, $z_{\text{Fundus}}, z_{\text{OCT}} \in \mathbb{R}^{(HW/P^2) \times D}$, are concatenated along the patch dimension as
\begin{equation}
    z_{\text{concat}} = \left[ z_{\text{Fundus}} ; z_{\text{OCT}} \right] \in \mathbb{R}^{2(HW/P^2) \times D}
\end{equation}
where $[ \cdot ; \cdot ]$ denotes the concatenation along the patch dimension.
This allows the Transformer to utilize the attention mechanism to consider the relationship between the two types of images.

On the other hand, integrating non-image data such as the patient's gender and age with image data can cause issues due to different input dimensions. Traditional approaches \cite{huang2020fusion} often use CNNs to incorporate medical record information into the channel dimension of input images, but these methods do not effectively utilize the complex relationships between medical record data and image features.
Therefore, we define $t$ as the number of types of medical record attributes, such as age, gender, and smoking history, and shape the medical record information into a vector. 

$T\in \mathbb{R}^{t}$. This vector is then transformed into table tokens $z_{table} \in \mathbb{R}^{T \times D}$ using a Multilayer Perceptron (MLP). These table tokens, matched in dimension to the image patch embeddings, enable the application of the attention mechanism with information from images.
The table embeddings, $z_{\text{table}} \in \mathbb{R}^{T \times D}$, are concatenated with the previously concatenated image embeddings $z_{\text{concat}}$ along the patch dimension as
\begin{equation}
    z_{\text{final}} = \left[ z_{\text{Fundus}} ; z_{\text{OCT}}; z_{\text{table}} \right] \in \mathbb{R}^{(2(HW/P^2) + T) \times D}
\end{equation}
where $[ \cdot ; \cdot ]$ denotes the concatenation along the patch dimension.

\subsubsection{Selective Transformer (ST)}

\begin{figure}[ht]
        \centering
        \includegraphics[width=0.6\linewidth]{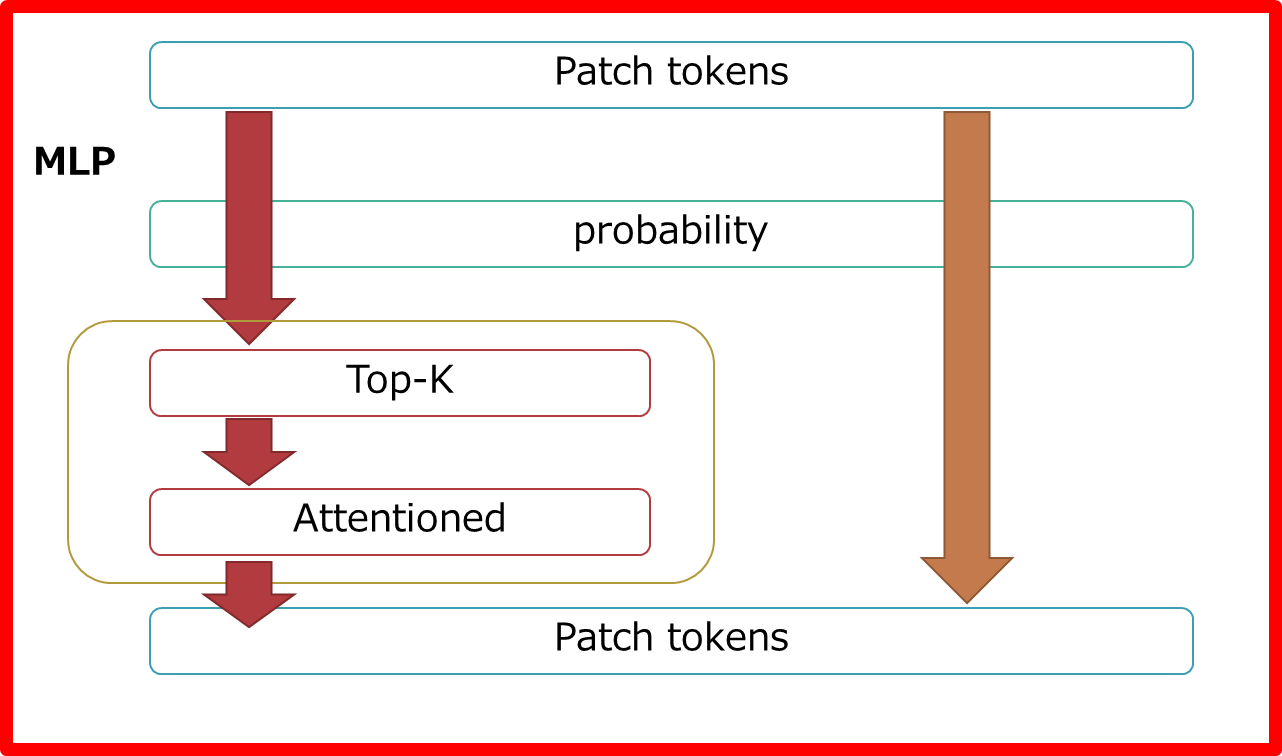}
        \caption{Selective Attention : Each image generates N tokens through embedding, which pass through an MLP to produce selection probabilities $P_{N}$. Only the tokens with the top K$\%$ probabilities are used for attention, resulting in more efficient processing.}
        \label{fig:Sel}
    \end{figure}

\begin{figure}[ht]
        \centering
        \includegraphics[width=0.8\linewidth]{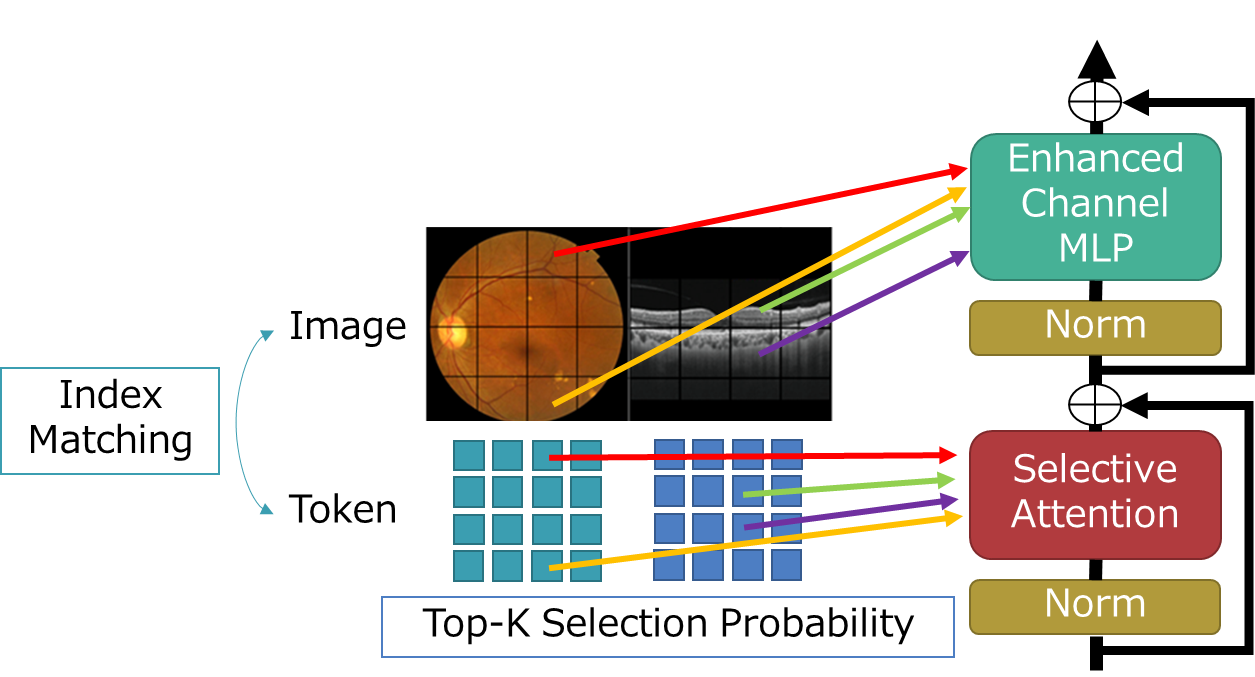}
        \caption{The overview of ST module}
        \label{fig:STall}
    \end{figure}

In medical imaging, the background portions of an image which are irrelevant to classification may be included, or the images under classification may be nearly identical, leading to potentially meaningless computational processes. 
Furthermore, when we handle multiple images simultaneously using the attention mechanism, treating all tokens derived from the images can be considered redundant. 
Therefore, we propose the Selective Transformer (ST), as illustrated in Figure \ref{fig:STall}, which introduces selective attention, shown in Figure \ref{fig:Sel}, to tokens based on learned probabilities and dense feature extraction using a CNN. First, each image generates N tokens through embedding. These tokens then pass through an MLP to output their respective selection probabilities $P_{N}$. The process can be described as
\begin{equation}
    \label{quad}
    P_N = \text{MLP}(z_N) \in \mathbb{R}^N
\end{equation}
where $z_N \in \mathbb{R}^{N \times D}$ represents the $N$ tokens generated from the image embedding, and $\text{MLP}(\cdot)$ denotes the Multi-Layer Perceptron. Since the tokens are directly transformed into probabilities without converting them to queries and keys, this approach is computationally less expensive and easier to implement than k-NN Attention \cite{knn}.

Only the tokens with the top K of selection probabilities are used to perform attention. This allows for more efficient attention application. This can be  expressed as
\begin{equation}
    \text{Top-}K(P_N) = { z_i \mid P_{N_i} \text{ is among the top } K \text{ probabilities} }
\end{equation}
where $\text{Top-}K(P_N)$ represents the selected tokens whose probabilities are among the top $K$.
The attention mechanism is then applied to these selected tokens, leading to more efficient attention application as
\begin{equation}
    \text{Attention}(Q_{s}, K_{s}, V_{s}) = \text{softmax}\left(\frac{Q_{s} K_{s}^T}{\sqrt{d_k}}\right)V_{s}
\end{equation}
where $Q_s$, $K_s$, and $V_s$ are the query, key, and value matrices derived from the selected tokens.
This selective attention mechanism ensures that only the most relevant tokens are processed, enhancing computational efficiency and potentially improving classification performance.

Furthermore, since the patch tokens used in attention can be considered as representing important regions, we introduce a more detailed feature extraction mechanism. These patch tokens are embedded from the original image using MME (Multi-Modal Embedding). As MME performs embedding on a patch-by-patch basis with patches of size p $\times$ p, it can be considered as aggregating global information. Therefore, we redefine $z_N$ in equation \ref{quad} as $z_{global}$.When we consider all image patches, the patch tokens are defined as
\begin{equation}
    z_{\text{global}} \in \mathbb{R}^{N \times D}
\end{equation}
The patch images are defined as
\begin{equation}
    I_{\text{patch}_i} \in \begin{cases}
    \mathbb{R}^{3 \times p \times p} & \text{for fundus images} \\
    \mathbb{R}^{1 \times p \times p} & \text{for OCT images}
    \end{cases}
    \quad \text{for} \quad i = 1, 2, \ldots, N
\end{equation}
The local feature representation $z_{local}$ is obtained as
\begin{equation}
z_{\text{local}_i} = \begin{cases}
    \text{CNN}(I_{\text{patch}_i}) & \text{if z}_i \text{ is selected} \\
    \text{Zeros}(I_{\text{patch}_i}) & \text{if z}_i \text{ is not selected}
    \end{cases}
    \quad \text{for} \quad i = 1, 2, \ldots, N
\end{equation}
where $z_{local}$ has the shape N $\times$ $D_{local}$ and $CNN(\cdot)$ is composed of 3 $\times$ 3 convolution, batch normalization, and ReLU function. The 3 $\times$ 3 convolution is denser than the convolution in the MME. The $Zeros(\cdot)$ produces a tensor of zeros with the same shape as the output of $CNN(\cdot)$. Subsequently, $z_{local}$ is concatenated with the output of the selective attention $z_{global}$ along the channel dimension.
\begin{equation}
    z_{\text{global,local}} = [z_{\text{global}} ; z_{\text{local}}] \in \mathbb{R}^{N \times (\text{D} + \text{D}_{\text{local}})}
\end{equation}
where $[ \cdot ; \cdot ]$ denotes the concatenation along the channel dimension.
These tokens are then processed through a channel MLP to integrate the information.
\begin{equation}
    z_{\text{integrated}} =  \text{Channel MLP}(z_{\text{global,local}},Zeros(z_{table}))
\end{equation}
where $z_{table}$ represents non-image information, and $Zeros(\cdot)$ the aligns its shape with $z_{global}$ and $z_{local}$. $Channel MLP(\cdot)$ represents the channel-wise Multi-Layer Perceptron that integrates the local features. This integration allows for recognizing important regions globally while considering the local information of those regions, thereby potentially improving the overall performance of the attention mechanism and the subsequent classification task.

\subsubsection{Enhanced head}

There are inherent size and shape differences between images and medical records, which can be considered as differences in the amount of information they contain. When we classify classes while simultaneously handling modalities with large differences in information content, there is a concern that the less informative medical records may not significantly influence the classification results, thus not contributing to improve the accuracy. Therefore, during training, in addition to standard classification, we perform a reconstruction of medical record information from the inputs to the classification head using Record-revive algorithm (RRA) module composed of a simple Multilayer Perceptron (MLP). 
Additionally, in this paper, we predict the risk alleles for two genes from a single set of inputs. The loss function used in training includes the cross-entropy loss for each of the ARMS2 and CFH genes, and the mean squared loss (MSE) for reconstructing the medical record information.
The total loss $L_{total}$ is defined as
\begin{equation}
    L_{total} = L_{CE_{ARMS2}} + L_{CE_{CFH}} + \alpha L_{MSE}
\end{equation}
where the cross-entropy loss for the ARMS2 gene is defined as
\begin{equation}
    L_{CE_{ARMS2}} = - \sum_{c=1}^{C} y_{c}^{ARMS2} \log(\hat{y}_{c}^{ARMS2})
\end{equation}
and the cross-entropy loss for the CFH gene is defined as
\begin{equation}
    L_{CE_{CFH}} = - \sum_{c=1}^{C} y_{c}^{CFH} \log(\hat{y}_{c}^{CFH})
\end{equation}
$C$ is the number of classes, $y_{c}^{ARMS2}$ and $y_{c}^{CFH}$ are the true labels for the ARMS2 and CFH genes respectively. 
$\hat{y}{c}^{ARMS2}$ and $\hat{y}{c}^{CFH}$ are the predicted probabilities for the ARMS2 and CFH genes respectively.
The mean squared error (MSE) loss for reconstructing the medical record information is defined as
\begin{equation}
    L_{MSE} = \frac{1}{t} \sum_{i=1}^{t} (x_i - \hat{x}_i)^2
\end{equation}
where $t$ is the number of medical record entries, $x_i$ is the true value, and $\hat{x}_i$ is the value predicted by RRA. The parameter $\alpha$ is a weighting factor that balances the contribution of the $L_{MSE}$ in the $L_{total}$.
This loss function
ensures that the model takes into account both the classification accuracy for the ARMS2 and CFH genes and the accuracy of reconstructing the medical record information, thereby potentially improving the overall performance of the model.

\subsection{Visualization of Selected Tokens}

In this paper, selective attention is used in the ST module. Therefore, traditional visualization techniques for Transformers that use all tokens, such as Attention Rollout \cite{chefer2021transformer}, cannot be directly employed. However, visualizing the rationale behind AI model decisions\cite{cam,gradcam,scorecam} is crucial in the medical field. Consequently, we consider tokens that are frequently selected by the ST module as important for predicting the number of genes. Thus, we use the frequency of selection for each 
token as a method of visualization.
Let $M$ be the number of blocks in the ST module. For each token $z_i$
\begin{equation}
    f_i = \sum_{m=1}^{M} \mathbb{I}(z_i \text{ is selected in block } m)
\end{equation}
where $f_i$ is the frequency of selection for token $z_i$, and $\mathbb{I}(\cdot)$ is an indicator function that returns 1 if the token $z_i$ is selected in block $m$, and 0 otherwise.
Each token can therefore have a frequency value $f_i$ ranging from 0 to $M$. The frequency of selection can be used as a method of visualization to highlight the importance of tokens in predicting the number of genes. Tokens with higher frequency values are considered more important.
\begin{equation}
    \text{Importance}(z_i) = f_i
\end{equation}
This approach allows for an effective visualization of tokens that ST module frequently selects, providing insight into the model's decision-making process.

\subsection{Table-based Similar Image Augmentation}

\begin{figure}[ht]
        \centering
        \includegraphics[width=0.8\linewidth]{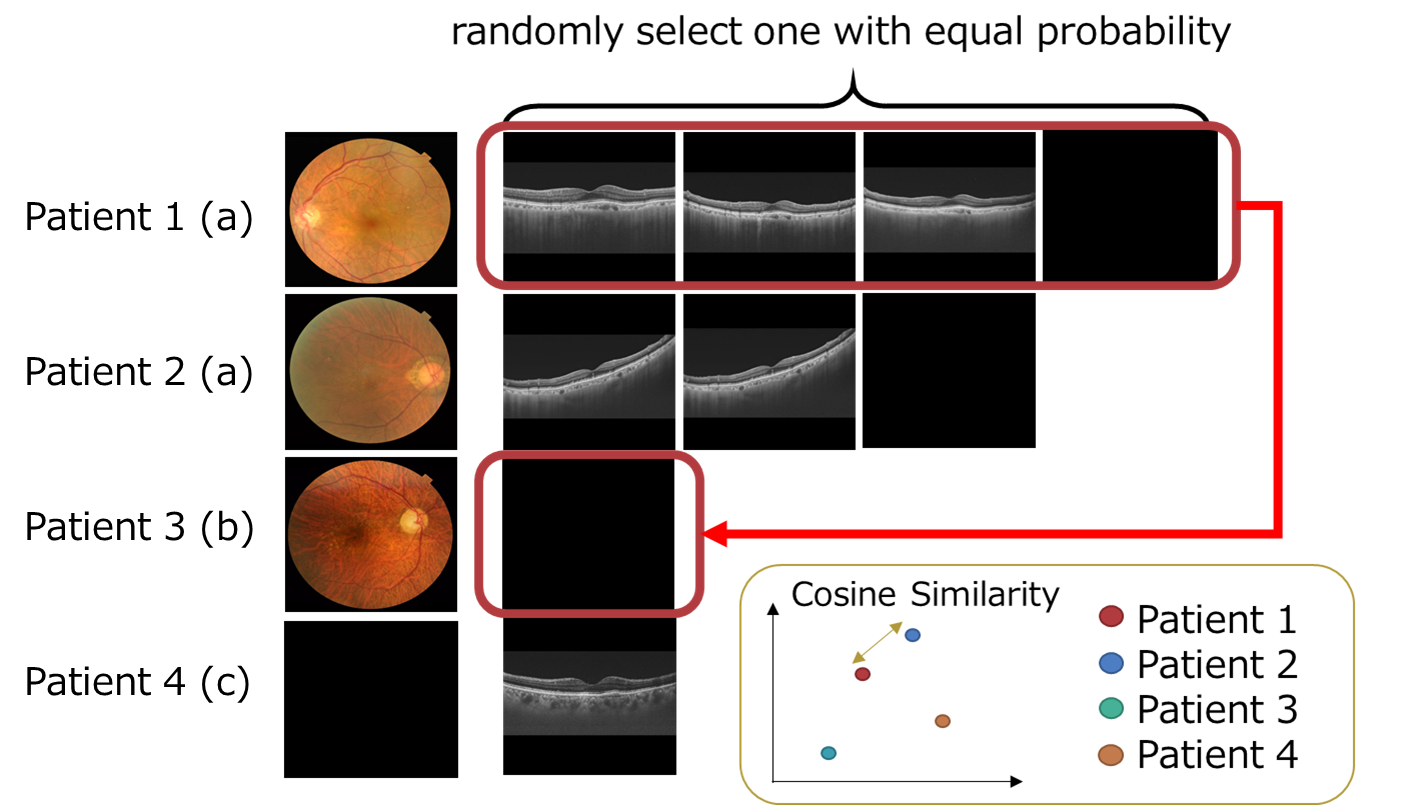}
        \caption{Table-based Similar Image Augmentation for OCT images}
        \label{fig:datagu}
    \end{figure}

For each patient, hospitals do not always capture both types of images (fundus and corresponding OCT images) during the same session, leading to different number of available fundus and OCT images. OCT images are captured when a physician deems a detailed examination of the eye necessary. Consequently, (a) multiple OCT images may correspond to a single fundus image, or (b) only one type of image, either a fundus or OCT image, may be available. Although the usage of a zero-value pseudo-image in place of the missing image type can solve the input shape issue for the model, extreme imbalances in availability of one type of image can lead to biased learning specific to image type. To address the issues, we propose a Table-based Similar Image Augmentation (TSIA) strategy. In (a), there are 1 to 3 OCT images per fundus image. But in the case (b), where no OCT images are available, it is necessary to train considering the absence of images. Therefore, OCT images are chosen at equal probability from pseudo-images and actual images with real values, as shown in Figure \ref{fig:datagu}. Furthermore, if no corresponding OCT images exist, a similar process is adopted by using the OCT image set of the patient with the same gene count and the highest cosine similarity in medical record information. This approach helps reduce the imbalance in the number of different types of images. Similarly, in the case (c), where the fundus image is missing, the same approach as in (b) can be applied.

\section{Experiments}
\subsection{Dataset}

The dataset used in this paper consists of 1,192 sets, each comprising a fundus image, a corresponding OCT image, a pseudo-image with all values set to zero, and medical records (age, gender, smoking history). While the dataset includes 1,172 fundus images, there are only 200 corresponding OCT images, which means only 200 sets include actual images. This indicates that the likelihood of each fundus image having a corresponding OCT image is very low. Each set is associated with the number of risk alleles (ranging from 0 to 2) for the ARMS2 and CFH genes. Cases with 0 or 1 risk allele are labeled as 0, and those with 2 risk alleles are labeled as 1.

Therefore, the task of this study is to classify whether the number of risk alleles for each gene is 2 based on the information derived from multiple images and medical records. Experiments will be conducted on sets of fundus images, OCT images, and medical records both with and without one type of image.

\subsection{Training and Evaluation Methods}

For training, fundus images are resized to 288 $\times$

288 pixels RGB images, OCT images are 288 $\times$ 288 pixels grayscale images, and the dimensionality of the medical record information is three. Data augmentation is limited to random horizontal flips. Training is conducted over 200 epochs with an initial learning rate of 0.001 and is optimized using cosine annealing\cite{loshchilov2016sgdr}. Furthermore, the hyperparameter $\alpha$ of the loss function is set to 0.001.

In the experiments, the data are split into five sets, with three used for training, one for validation, and one for testing, implementing five-fold cross-validation. The evaluation metrics include accuracy, precision, recall, specificity, and the F-score, which is the harmonic mean of precision and recall.

\subsection{Results}

The prediction results for the ARMS2 gene are shown in Table \ref{tab:ARMS2}, and the results for the CFH gene are presented in Table \ref{tab:CFH}. When we compared to the scenarios without OCT images, the proposed method improved the accuracy by approximately 7\% and the F-score by about 21\% for the ARMS2 gene. For the CFH gene, accuracy improved by about 15\% and the F-score by approximately 12\%. Furthermore, when we compared to the scenarios without fundus images, the accuracy for the ARMS2 gene is improved by about 10\% and the F-score by about 38\%. For the CFH gene, the accuracy is improved by approximately 16\% and the F-score by about 16\%.
    \begin{table}[ht]
        \begin{center}
        \caption{Results for the ARMS2 Gene}
        \label{tab:ARMS2}
        \begin{tabular}{|l|c|c|c|c|c|}
        \hline
        Method & Accuracy & Precision & Recall & Specificity & F-score\\
        \hline\hline
        Without OCT & 76.42\% & 60.98\% & 42.37\% & 94.14\% & 50.00\% \\
        Without Fundus & 73.84\% & 86.00\% & 20.66\% & 98.36\% & 33.31\% \\
        Proposed Method & 83.44\% & 81.06\% & 62.96\% & 94.00\% & 70.87\% \\
        \hline
        \end{tabular}
        \end{center}
    \end{table}
    \begin{table}[ht]
        \begin{center}
        \caption{Results for the CFH Gene}
        \label{tab:CFH}
        \begin{tabular}{|l|c|c|c|c|c|}
        \hline
        Method & Accuracy & Precision & Recall & Specificity & F-score\\
        \hline\hline
        Without OCT & 65.74\% & 68.22\% & 71.98\% & 59.58\% & 70.05\% \\
        Without Fundus & 64.65\% & 68.26\% & 64.43\% & 65.15\% & 66.29\% \\
        Proposed Method & 80.66\% & 79.56\% & 84.97\% & 76.01\% & 82.18\% \\
        \hline
        \end{tabular}
        \end{center}
    \end{table}

\subsection{Visualization of Selected Tokens}
\begin{figure}[ht]
        \centering
        \includegraphics[width=0.7\linewidth]{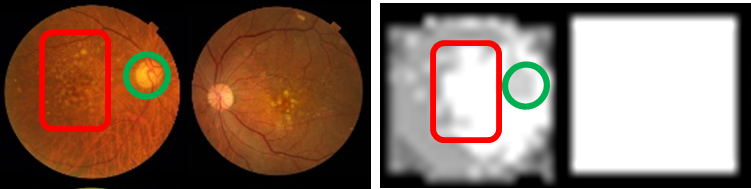}
        \caption{Areas like optic disc (green outline), less relevant to AMD, have lower selection frequencies, while drusen (red outline) and surrounding areas show higher frequencies.}
        \label{fig:f}
    \end{figure}

\begin{figure}[ht]
        \centering
        \includegraphics[width=0.7\linewidth]{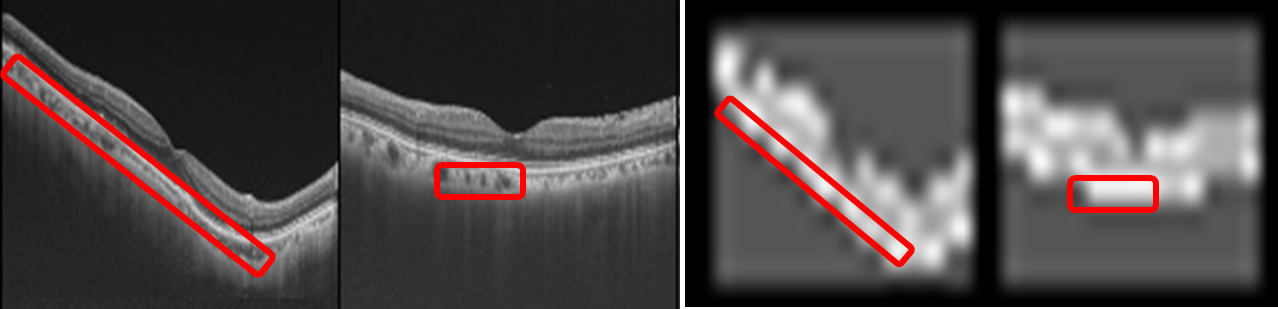}
        \caption{The choroid (red outline) in the OCT image also shows high selection frequency, indicating the ST module focuses on areas relevant to AMD susceptibility genes.}
        \label{fig:o}
    \end{figure}

We present the visualization of tokens probabilistically selected by the ST module, where tokens with higher selection frequencies appear whiter in Figures \ref{fig:f} and \ref{fig:o}. In the case where OCT images are absent from the dataset, as shown in Figure \ref{fig:f}, tokens derived from fundus information are selected, resulting in an entirely white image. Furthermore, areas such as the optic disc (outlined in green), which are considered less relevant to AMD susceptibility genes, show lower selection frequencies. In contrast, tokens containing drusen (outlined in red) in the fundus image and their surrounding areas exhibit higher selection frequencies. Additionally, in the OCT image shown in Figure \ref{fig:o}, the choroid (outlined in red) shows a high selection frequency. This suggests that the ST module learns to select tokens surrounding areas highly relevant to AMD susceptibility genes.
    
\section{Ablation Study}
\subsection{Impact of TSIA on Accuracy}
\begin{figure}[htb]
        \centering
        \includegraphics[width=0.8\linewidth]{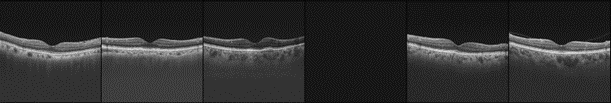}
        \caption{Input OCT images}
        \label{fig:octs}
    \end{figure}

\begin{figure}[htb]
        \centering
        \includegraphics[width=0.8\linewidth]{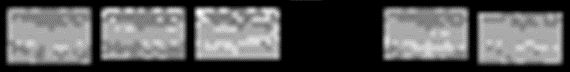}
        \caption{In the case where TSIA was not used, tokens corresponding to background noise present in the OCT images were also selected to some extent.}
        \label{fig:s0octs}
    \end{figure}

\begin{figure}[htb]
        \centering
        \includegraphics[width=0.8\linewidth]{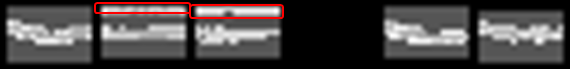}
        \caption{Applying TSIA results in the model learning to preferentially select tokens that correspond to the cross section rather than the background noise.}
        \label{fig:s1octs}
    \end{figure}

\begin{figure}[htb]
        \centering
        \includegraphics[width=0.9\linewidth]{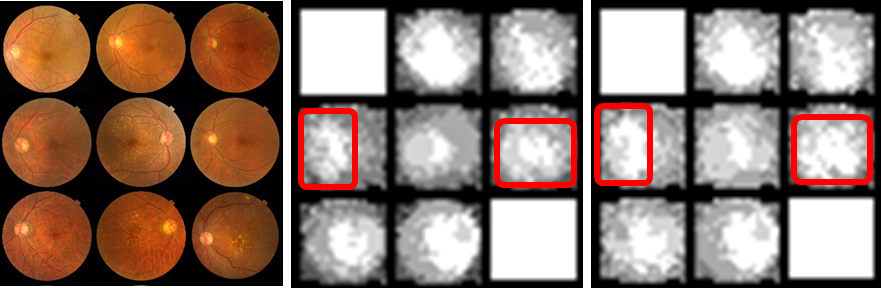}
        \caption{Since background noise in the OCT images was rarely selected, tokens corresponding to the fundus were chosen instead.}
        \label{fig:funds}
    \end{figure}

This study implemented TSIA to address the unmatch in the number of fundus and OCT images. Tables \ref{tab:augA} and \ref{tab:augC} compare the accuracy with and without the TSIA. It is evident from these Tables that particularly the recall rates improved with the application of TSIA. As can be seen in Tables \ref{tab:augA} and \ref{tab:augC}, OCT images contribute to the improvement in recall. Therefore, it can be inferred that augmenting OCT images through TSIA not only enhances recall but also contributes to overall accuracy improvement.
Furthermore, the changes in the selected tokens with and without TSIA are illustrated in the Figure \ref{fig:octs} to \ref{fig:funds}.
As can be seen from Figure \ref{fig:s0octs}, in the case where TSIA was not applied, tokens corresponding to background noise present in the OCT images were also selected to some extent. In contrast, the Figure \ref{fig:s1octs} shows that applying TSIA results in the model learning to preferentially select tokens corresponding to the cross section rather than the background noise.
Additionally, as seen in the fundus images in Figure \ref{fig:funds}, since background noise in the OCT images was hardly selected, it is evident that tokens corresponding to the fundus in the fundus images are being selected instead when we handled simultaneously with OCT images. These results indicate that TSIA enables the model to selectively remove background noise and choose the fundus necessary for classification by pseudo-learning the missing OCT images of the patients.
    
    \begin{table}[ht]
        \begin{center}
        \caption{Changes in accuracy due to TSIA for ARMS2}
        \label{tab:augA}
        \begin{tabular}{|l|c|c|c|c|c|}
        \hline
        Method & Accuracy & Precision & Recall & Specificity & F-score\\
        \hline\hline
        Without TSIA & 82.05\% & 80.94\% & 54.52\% & 94.71\% & 65.16\% \\
        With TSIA & 83.44\% & 81.06\% & 62.96\% & 94.00\% & 70.87\% \\
        \hline
        \end{tabular}
        \end{center}
    \end{table}
    \begin{table}[ht]
        \begin{center}
        \caption{Changes in accuracy due to TSIA for CFH}
        \label{tab:augC}
        \begin{tabular}{|l|c|c|c|c|c|}
        \hline
        Method & Accuracy & Precision & Recall & Specificity & F-score\\
        \hline\hline
        Without TSIA & 78.21\% & 77.47\% & 82.01\% & 74.31\% & 79.68\% \\
        With TSIA & 80.66\% & 79.56\% & 84.97\% & 76.01\% & 82.18\% \\
        \hline
        \end{tabular}
        \end{center}
    \end{table}

\subsection{Impact of Medical Record Information on Accuracy}

In this paper, not only images but also medical record information were used. Tables \ref{tab:infoA} and \ref{tab:infoC} show a comparison between cases where medical record information is available, not available, and cases where training is conducted to reconstruct the medical record information when it is available. For the ARMS2 gene, merely including medical record information improved the accuracy by approximately 2\% and the F-score by about 14\%. However, for the CFH gene, including medical record information alone resulted in an increase of approximately 8\% in accuracy and a 9\% improvement in the F-score. This may be attributed to the relatively smaller amount of information contained in medical records compared to images, which reduces their impact on classification. Nonetheless, when medical record information was reconstructed, the accuracy improved by approximately 2\% and the F-score by about 7\% for the ARMS2 gene. For the CFH gene, the accuracy improved by approximately 3\% and the F-score by about 4\%. These results indicate that it is crucial to simultaneously perform the reconstruction of medical record information when incorporating it with images.

    \begin{table}[H]
        \begin{center}
        \caption{Necessity of Medical Record Information for ARMS2. For the ARMS2 gene, adding medical records improved the accuracy by 2\% and the F-score by 14\%}
        \label{tab:infoA}
        \begin{tabular}{|l|c|c|c|c|c|}
        \hline
        Method & Accuracy & Precision & Recall & Specificity & F-score\\
        \hline\hline
        Without Information & 79.61\% & 66.67\% & 40.74\% & 96.82\% & 50.57\% \\
        With Information & 81.97\% & 77.61\% & 54.94\% & 95.06\% & 64.33\%\\
        + Reconstruction & 83.44\% & 81.06\% & 62.96\% & 94.00\% & 70.87\% \\
        \hline
        \end{tabular}
        \end{center}
    \end{table}
    
    \begin{table}[H]
        \begin{center}
        \caption{Necessity of Medical Record Information for CFH. For the CFH gene, the accuracy decreased by 8\% and the F-score by 9\%. }
        \label{tab:infoC}
        \begin{tabular}{|l|c|c|c|c|c|}
        \hline
        Method & Accuracy & Precision & Recall & Specificity & F-score\\
        \hline\hline
        Without Information & 69.53\% & 74.37\% & 65.62\% & 73.88\% & 69.72\% \\
        With Information & 77.51\% & 78.43\% & 78.17\% & 77.15\% & 78.30\%\\
        + Reconstruction & 80.66\% & 79.56\% & 84.97\% & 76.01\% & 82.18\% \\
        \hline
        \end{tabular}
        \end{center}
    \end{table}

\subsection{Token Selection Rate}
\begin{figure}[htb]
        \centering
        \includegraphics[width=0.65\linewidth]{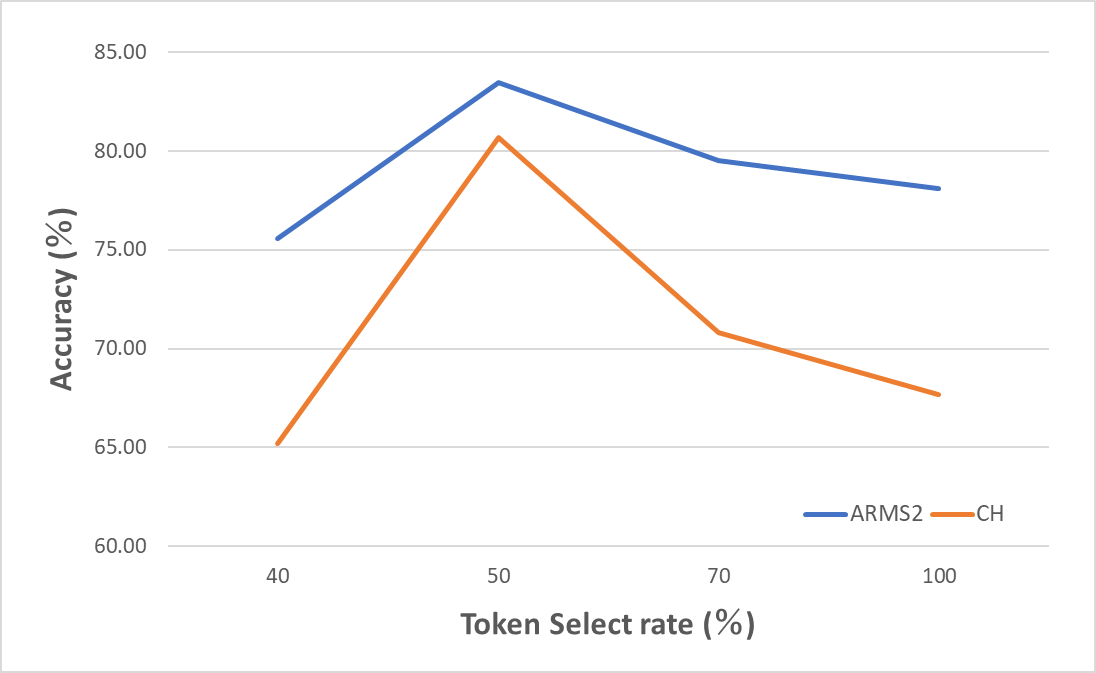}
        \caption{Change in accuracy depending on selection rate. The highest accuracy is achieved at a 50\% token selection rate. This suggests that the relevant tokens at a lower selection rate enhances the accuracy by capturing essential information, while rates below 50\% reduce image information too much, decreasing the accuracy. A 100\% selection rate corresponds to the standard Multi-Head Attention.}
        \label{fig:selective}
    \end{figure}

In this paper, the ST module is utilized to select tokens for applying attention. The accuracy improvements for the ARMS2 and CFH genes with and without the ST module are demonstrated in Tables \ref{tab:ARMS2ST} and \ref{tab:CFHST}. With the ST module, the accuracy for the ARMS2 gene was improved by approximately 5\%, and the F-score by about 20\%. For the CFH gene, the accuracy was improved by approximately 13\%, and the F-score by about 15\%. Additionally, changes in accuracy are shown when varying the token selection rate.
From Figure \ref{fig:selective}, Table \ref{tab:ARMS2ST}, and Table \ref{tab:CFHST}, it is evident that the presence of the ST module leads to higher accuracy. For this dataset, the highest accuracy occurs at a 50\% selection rate. This suggests that a lower selection rate focuses attention on the most relevant tokens, enhancing the accuracy by capturing global information while increasing locality. However, when the selection rate drops below 50\%, it is presumed that excessive reduction of image information occurs, leading to decrease the classification accuracy.

    \begin{table}[H]
        \begin{center}
        \caption{ST module improved 5\% for ARMS2 gene and 20\% for F-score}
        \label{tab:ARMS2ST}
        \begin{tabular}{|l|c|c|c|c|c|}
        \hline
        Method & Accuracy & Precision & Recall & Specificity & F-score\\
        \hline\hline
        Without ST & 78.11\% & 74.53\% & 39.06\% & 95.72\% & 51.26\% \\
        With ST & 83.44\% & 81.06\% & 62.96\% & 94.00\% & 70.87\% \\
        \hline
        \end{tabular}
        \end{center}
    \end{table}
    
    \begin{table}[H]
        \begin{center}
        \caption{ST module improved 13\% for CFH gene 15\% for F-score}
        \label{tab:CFHST}
        \begin{tabular}{|l|c|c|c|c|c|}
        \hline
        Method & Accuracy & Precision & Recall & Specificity & F-score\\
        \hline\hline
        Without ST & 67.69\% & 72.78\% & 61.64\% & 74.79\% & 66.75\% \\
        With ST & 80.66\% & 79.56\% & 84.97\% & 76.01\% & 82.18\% \\
        \hline
        \end{tabular}
        \end{center}
    \end{table}
    
\FloatBarrier

\section{Conclusions}

In this study, we conducted experiments to identify multiple genes related to diseases based on the number of bases, integrating multiple images with different imaging methods and color channels, as well as medical record information, to closely resemble a physician's diagnosis. 
The prediction results demonstrated that effectively integrating multi-modality information can predict AMD susceptibility genes with high performance. Particularly, the reconstruction of medical records, the use of the ST module, and the TSIA significantly contributed to the high performance. 

For both ARMS2 and CFH genes, simultaneously handling multiple images and Table information significantly improved the accuracy and F-score. When the selection rate decreases, the accuracy increased, indicating the effectiveness of the ST module's selective attention and detailed feature extraction. This improvement is supported by the increased accuracy and is further evidenced by the visualization of the selected tokens.
Moreover, the TSIA module improved the selection by addressing missing OCT images through the use of similar fundus images, thus handling image data with missing parts. The selection rate is influenced by the high amount of background noise in OCT images. The accuracy improved as the selection rate approaches half of the patch tokens due to the significant background noise in OCT images. It would be beneficial to determine the selection rate through learning to enhance this aspect further.




\newpage
%
%
\bibliographystyle{splncs04}
\bibliography{main}
\end{document}